\documentclass[10pt,twocolumn,letterpaper]{article}
\usepackage[T1]{fontenc}
\usepackage[utf8]{inputenc}
\usepackage[margin=0.75in,columnsep=0.25in]{geometry}

\usepackage{newtxtext}
\usepackage{amsmath,amsfonts,amssymb}
\usepackage{newtxmath}

\usepackage{amsmath,amsfonts,bm}

\def\eqref#1{equation~\ref{#1}}

\def\1{\bm{1}}

\DeclareMathAlphabet{\mathsfit}{\encodingdefault}{\sfdefault}{m}{sl}
\SetMathAlphabet{\mathsfit}{bold}{\encodingdefault}{\sfdefault}{bx}{n}

\usepackage{graphicx,booktabs,tabularx,makecell,multirow,capt-of}
\usepackage{listings,xcolor,enumitem,xspace,float}
\usepackage[numbers,sort&compress]{natbib}
\usepackage{caption}
\usepackage{subcaption}
\usepackage{xurl}
\usepackage[hidelinks]{hyperref}

\definecolor{codegray}{RGB}{245,245,245}

\newcommand{\sysname}{\textsc{JS$^2$}\xspace}
\title{\sysname: Joint Scheduling and Speculative Decoding\\for Efficient Agentic LLM Inference}

\author{
    \normalsize
    Zijian Wen$^{1}$,
    Liangkun Chen$^{1}$,
    Jiamin Cao$^{2}$,
    Ennan Zhai$^{2}$,
    Songjun Huang$^{1}$,
    Xu Chen$^{1}$,
    Xiaoxi Zhang$^{1,*}$\\[5pt]
    \small
    $^{1}$Sun Yat-sen University
    \qquad
    $^{2}$Alibaba Group
}

\date{}

\begin{document}
\twocolumn[
\begin{@twocolumnfalse}
\maketitle
\begin{abstract}
LLM-powered agents execute tasks through interdependent LLM calls interleaved with tool executions, making end-to-end Job Completion Time (JCT) a central serving objective.
Our measurements reveal that the dominant LLM inference bottleneck shifts from decoding latency at low concurrency to queueing delay at high concurrency.
Speculative Decoding (SD) accelerates decoding but may degrade performance under heavy load.
Dynamic arrivals and intermittent tool executions cause batch-size fluctuations in online agent serving.
Existing Adaptive Speculative Decoding (ASD) approaches rely on reactive feedback, incur overhead when speculation is disabled, and offer limited upper-layer control.
Agent-level scheduling mitigates queueing delays but overlooks the load-dependent effectiveness of SD.
Our key insight is that prioritizing tasks with fewer remaining LLM requests reduces queueing delays and shortens high-batch periods, creating additional SD opportunities.
We present \sysname, a bottleneck-aware LLM serving system that jointly optimizes SD and agent-level scheduling.
\sysname enables externally controllable, iteration-level switching between native SD and non-SD paths using separate CUDA Graphs, minimizing disabled-SD overhead while preserving ASD compatibility.
It estimates remaining steps using explicit plans, prompt-derived execution states, and hierarchical matching of historical trajectories.
\sysname coordinates remaining-step-aware scheduling with load-adaptive SD selection based on offline profiling and near-future load estimation.
We implement \sysname on vLLM and SGLang and evaluate it across three LLMs, four workloads, three workflows, two GPU platforms, and two real-world serving traces.
\sysname achieves up to a 2.10$\times$ speedup in average JCT over the best baselines, while maintaining comparable tail latency and throughput.
Our code is available at \url{https://github.com/Wenz0101/JS2}.
\end{abstract}
\vspace{1em}
\end{@twocolumnfalse}
]

\begingroup
\renewcommand{\thefootnote}{\fnsymbol{footnote}}
\footnotetext[1]{Corresponding author: Xiaoxi Zhang}
\endgroup

\section{Introduction}

\begin{figure*}[!t]
    \centering
    \includegraphics[width=0.95\linewidth]{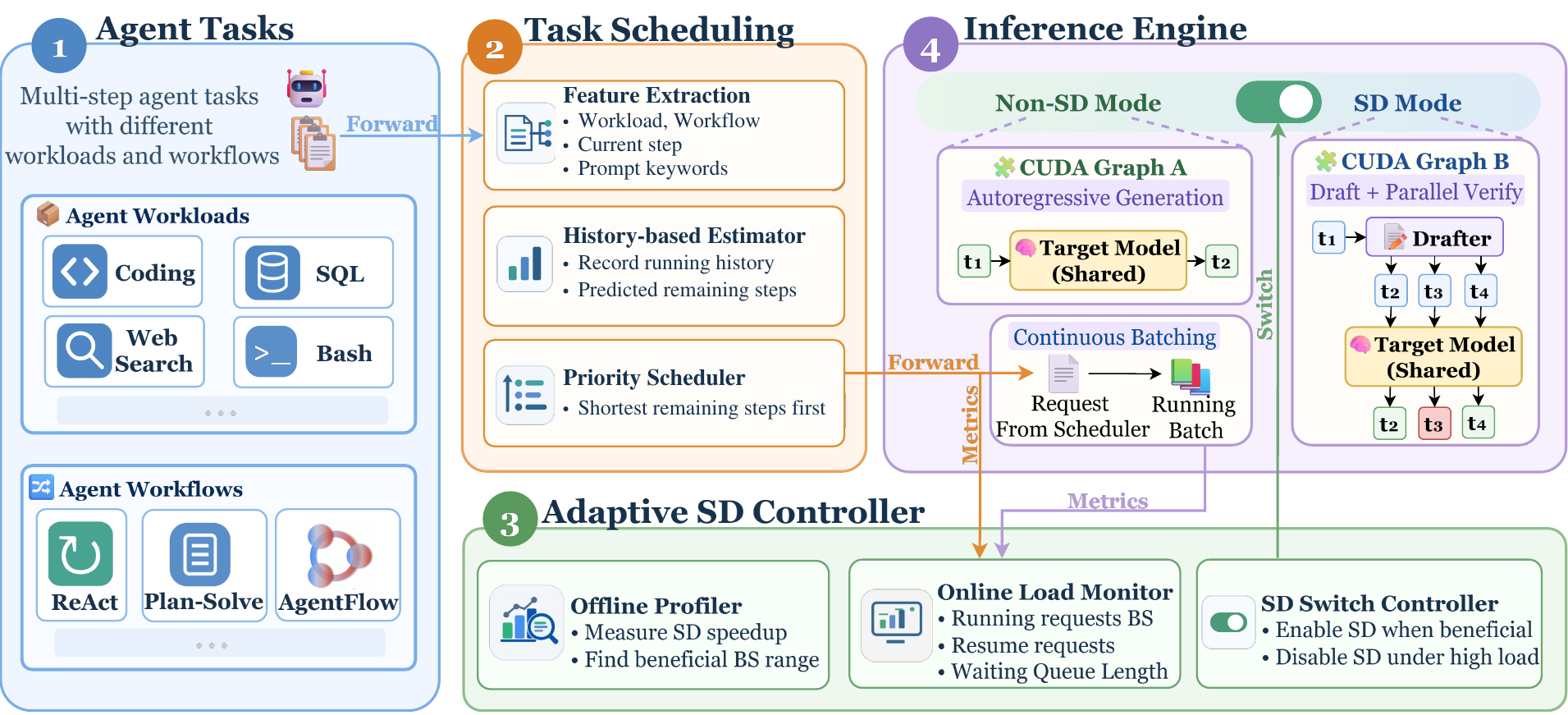}
    \caption{
    System overview of \sysname, combining remaining-step-aware agent scheduling with load-adaptive SD switching.
    }
    \label{fig:overview}
\end{figure*}

Large language models (LLMs) have evolved from chatbots into autonomous agents capable of solving complex tasks, such as programming~\citep{MBPP}, web search~\citep{WebGPT}, and environment interaction~\citep{InterCode}.
Unlike conventional LLM requests, representative agentic workflows~\citep{ReAct,Plan-and-Solve,AgentFlow} involve multiple interdependent LLM calls interleaved with tool executions.
Consequently, users care about the end-to-end Job Completion Time (JCT) of the entire agent task rather than the latency of individual LLM requests, posing significant challenges for existing LLM serving systems.
LLM inference accounts for the majority of end-to-end agent latency (see~\autoref{llm-time-ratio}).
Our measurements (\autoref{fig:motivation-a}) reveal that the dominant performance bottleneck shifts from decoding latency at low concurrency to queueing delay at high concurrency.

Speculative Decoding (SD) accelerates LLM inference by generating candidate tokens with a lightweight drafting mechanism and verifying them in parallel using the target model~\citep{SD1,SD2,Eagle}.
Agent workloads provide additional opportunities for SD due to their repetitive and structured generation patterns, such as recurring intermediate outputs, tool calls, and schema-constrained responses~\citep{SuffixDecoding,ToolSpec,AgentSpec}.
However, SD speedup decreases as the running batch size increases, potentially degrading performance under high load~\citep{SD-Speedup,SSSD,TurboSpec}.
This limitation is particularly important in online agent serving, where dynamic request arrivals and intermittent tool executions cause substantial fluctuations in running batch size.

Existing Adaptive Speculative Decoding (ASD) approaches dynamically adjust speculation based on runtime signals such as batch size and token acceptance rate~\citep{TurboSpec,SpecDec++,Nightjar}.
However, engine-level ASD implementations typically rely on reactive runtime feedback and retain substantial overhead even when speculation is disabled, limiting adaptation flexibility and inference efficiency.
More fundamentally, ASD passively adapts to serving load without directly addressing the dominant queueing bottleneck under high concurrency.

Meanwhile, existing LLM schedulers prioritize individual requests~\citep{SSJF,TRAIL,FastServe}, while agent-aware serving systems optimize task scheduling and resource management~\citep{Agentix,ThunderAgent}.
Although scheduling can mitigate queueing delays, it does not directly accelerate decoding or explicitly exploit the load-dependent effectiveness of SD.
Our key insight is that prioritizing agent tasks with fewer remaining LLM requests can shorten high-batch periods, allowing the system to enter the SD-beneficial range earlier (\autoref{fig:motivation-b}).
Therefore, scheduling not only reduces queueing delays but also creates additional opportunities for SD acceleration.
This motivates jointly optimizing agent scheduling and SD rather than treating them as independent techniques.

To this end, we propose \sysname, a bottleneck-aware LLM serving system that jointly optimizes SD and agent-level scheduling.
As illustrated in~\autoref{fig:overview}, \sysname comprises four components.
\textbf{Agent Tasks} represent multi-step agent executions across diverse workloads and workflows.
\textbf{Task Scheduling} estimates remaining agent steps using execution history and prompt-derived states and prioritizes shorter tasks.
The \textbf{Adaptive SD Controller} combines offline profiling with online serving load estimation to adaptively select SD.
The \textbf{Inference Engine} supports externally controlled, iteration-level switching between native SD and non-SD execution paths while remaining compatible with existing ASD policies.
Together, these components address shifting performance bottlenecks and exploit the synergy between scheduling and SD.

Our main contributions are summarized as follows:

\noindent\textbf{Externally Controllable SD Switching.}
We design an iteration-level switching interface that enables external control over native SD and non-SD execution paths while remaining compatible with existing ASD policies.
This reduces non-SD overhead from 4.9\%-34.9\% to only 0.6\%-3.7\%.

\noindent\textbf{Lightweight Agent Remaining-Step Prediction.}
We develop a unified history-based estimator that predicts remaining agent steps using prompt-derived execution states and historical patterns, achieving 84.1\% pairwise ordering accuracy and 82.2\% top-10\% recall across diverse agent workloads.

\noindent\textbf{Joint Agent Scheduling and SD Optimization.}
We coordinate remaining-step-aware agent scheduling with load-adaptive SD selection.
By prioritizing tasks with fewer future LLM requests, \sysname mitigates queueing delays and creates additional SD acceleration opportunities, jointly addressing queueing and decoding bottlenecks.

Extensive evaluations across three LLMs, four agent workloads, three workflows, two GPU platforms, and two real-world online serving traces demonstrate that \sysname achieves up to a 2.10$\times$ speedup in average JCT under resource-constrained settings.

\section{Background and Motivation}

\subsection{Adaptive Speculative Decoding}
\label{sec:asd}

\begin{table}[!b]
    \centering
    \caption{Execution path support and overhead relative to native non-SD execution across adaptive SD approaches.}
    \label{tab:adaptive_sd_comparison}
    \resizebox{\columnwidth}{!}{
    \begin{tabular}{lccc}
        \toprule
        \textbf{Approach} &
        \textbf{non-SD Path} &
        \textbf{SD Path} &
        \textbf{Overhead} \\
        \midrule

        vLLM-DSD
        & $\times$ & $\checkmark$ & 13.7\%-34.9\% \\

        SGLang-ASD
        & $\times$ & $\checkmark$ & 4.9\%-13.2\% \\

        \sysname
        & $\checkmark$ & $\checkmark$ & 0.6\%-3.7\% \\

        \bottomrule
    \end{tabular}
    }
\end{table}

Speculative Decoding (SD) accelerates autoregressive LLM inference by generating candidate tokens with a lightweight draft model and verifying them in parallel using the target model.
However, its effectiveness depends on both token acceptance rate and serving load.
As batch size increases, the speedup of SD decreases and may eventually turn into performance degradation~\citep{SD-Speedup}.
\autoref{sd-speedup} provides detailed measurements across hardware platforms, model configurations, and draft lengths.

To address this limitation, existing Adaptive Speculative Decoding (ASD) approaches mainly explore two directions.
One direction adjusts drafting parameters, particularly draft length, based on runtime signals such as token acceptance rate, batch size, system load, and SLOs~\citep{SpecDec++,TurboSpec,BanditSpec,AdaSpec,Nightjar,Learning_To_Draft}.
Another direction optimizes verification by dynamically allocating or pruning verification workloads across requests~\citep{AdaServe,D-Cut,ECHO,DSpark}.
Similar adaptation mechanisms are implemented in vLLM's Dynamic SD~\citep{vLLM} and SGLang's Adaptive SD~\citep{SGLang}.

However, existing engine-level ASD implementations have two limitations.
First, their adaptation logic is tightly coupled with the inference engine and typically configured at startup, limiting the flexibility of upper-layer control.
Moreover, switching decisions primarily rely on reactive runtime feedback, such as current running batch size and historical token acceptance rate.
Such feedback-driven adaptation has limited foresight and may lag behind rapidly changing serving conditions, particularly in agent workloads with dynamic arrivals and intermittent tool executions.
Without an external control interface, upper-layer schedulers cannot proactively adjust SD based on workload information and scheduling decisions.
Second, existing implementations are confined to the SD execution path rather than supporting direct switching between native SD and non-SD paths.
Even when the draft length is reduced to zero, SD-specific execution logic still remains, as evidenced by the detailed token consistency analysis in~\autoref{execution-path-verification}, introducing non-negligible overhead compared with native non-SD execution.

As shown in~\autoref{tab:adaptive_sd_comparison}, vLLM-DSD and SGLang-ASD incur 13.7\%-34.9\% and 4.9\%-13.2\% non-SD overhead, respectively, compared to only 0.6\%-3.7\% for \sysname.
These limitations motivate an externally controllable, low-overhead switching mechanism between native SD and non-SD execution paths.

\subsection{Agent Task Execution Prediction}
\label{sec:task_prediction}

Although agent tasks exhibit dynamic execution patterns, their future execution demands are not entirely unpredictable.
Recent studies have exploited workflow characteristics, historical execution patterns, and runtime information to predict future computational demands for efficient agent serving~\citep{Pythia,Chimera,TokenCast}.
These studies suggest that agent execution patterns provide useful information for anticipating future resource demands.
Inspired by these observations, we investigate the predictability of agent execution steps, which directly reflect future LLM request demands.

\begin{figure}[!b]
    \centering
    \includegraphics[width=\columnwidth]{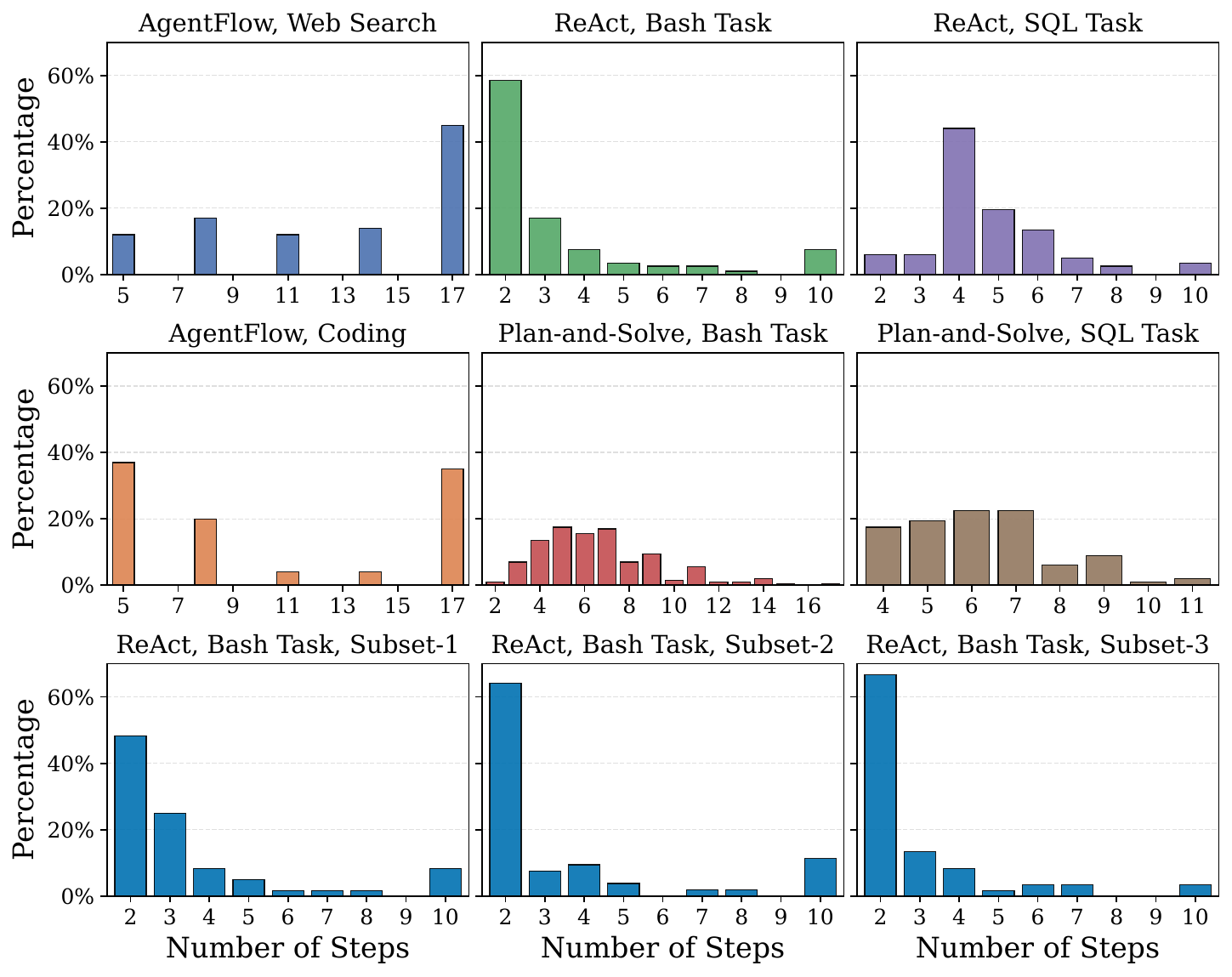}
    \caption{Step distributions across workflows and workloads.}
    \label{fig:step_predictability}
\end{figure}

To investigate the predictability of agent execution steps, we measure the distributions of total execution steps across different workflows and workloads.
As shown in the first two rows of~\autoref{fig:step_predictability}, step distributions differ substantially across workloads under the same workflow, as well as across workflows under the same workload.
In contrast, the last row shows that three different subsets under the same workflow--workload setting exhibit similar distributions.
These observations suggest that agent execution steps are strongly associated with workflow and workload and exhibit statistical regularities within the same setting.
This motivates constructing workflow- and workload-specific step distributions from historical execution records to estimate the execution steps of future agent tasks.

Beyond historical execution patterns, the remaining steps of an agent task also depend on its current execution state.
Intuitively, a failed execution step often requires additional retries or corrective actions compared with a successful one, resulting in more remaining steps.
Such execution states are typically reflected in the agent's interaction context through signals such as errors, retries, completion, and termination.
This motivates incorporating prompt-derived execution states into historical step estimation to refine remaining-step predictions at runtime.
\subsection{Agentic LLM Serving}
\label{sec:agent_serving}

LLM agents execute tasks through interdependent LLM calls interleaved with tool executions, making end-to-end Job Completion Time (JCT) a primary performance metric.
Existing LLM schedulers improve serving efficiency through predicted output lengths or relative length rankings~\citep{SSJF,LTR}, online remaining length estimation~\citep{TRAIL}, preemptive scheduling~\citep{FastServe}, and estimated future resource demands~\citep{Past-Future,JITServe}.
However, these methods primarily operate at the granularity of individual LLM requests, without explicitly considering subsequent agent steps.
Prioritizing individual LLM requests does not necessarily align with minimizing the end-to-end JCT of entire agent tasks.

Recent agent-aware serving systems further exploit application dependencies~\citep{Parrot}, attained service~\citep{Agentix}, historical execution states and predicted future stages~\citep{Astraea}, program-level execution and resource states~\citep{ThunderAgent}, and workflow structure and cache information~\citep{Helium}.
These approaches improve agent serving efficiency by incorporating upper-layer execution information into scheduling decisions.
However, their scheduling policies do not explicitly account for the batch-size sensitivity of SD.
Moreover, while scheduling can mitigate queueing delays, it does not directly accelerate token generation, limiting its effectiveness when decoding becomes the dominant bottleneck.

\begin{figure}[!tbp]
    \centering
    \begin{subfigure}[t]{0.40\columnwidth}
        \includegraphics[width=\linewidth]{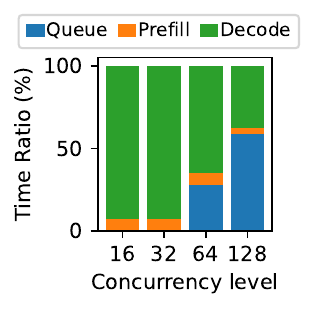}
        \caption{LLM bottleneck analysis in agent tasks.}
        \label{fig:motivation-a}
    \end{subfigure}
    \hfill
    \begin{subfigure}[t]{0.56\columnwidth}
        \includegraphics[width=\linewidth]{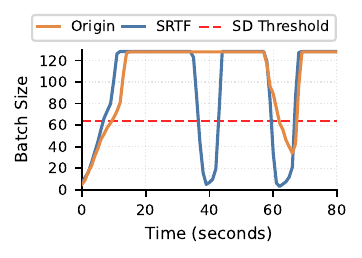}
        \caption{Running batch size under different scheduling policies.}
        \label{fig:motivation-b}
    \end{subfigure}
    \caption{Performance bottlenecks and scheduling opportunities in agentic LLM serving.}
    \label{fig:motivation}
\end{figure}

As shown in~\autoref{fig:motivation-a}, the dominant bottleneck shifts from decoding to queueing as concurrency increases.
This motivates using SD to accelerate decoding under low load and scheduling to reduce queueing delays under high load.
However, independently applying these techniques overlooks the impact of scheduling decisions on the effectiveness of SD.

As shown in~\autoref{fig:motivation-b}, intermittent tool executions cause sawtooth-like fluctuations in running batch size.
Prioritizing shorter agent tasks, as illustrated by Shortest Remaining Time First (SRTF) scheduling, completes them earlier and shortens subsequent high-batch periods.
This allows the running batch size to fall sooner into the SD-beneficial range (below the SD Threshold), creating additional acceleration opportunities.
Therefore, agent-level scheduling not only mitigates queueing delays but also creates favorable conditions for SD, which can further accelerate the decoding of remaining tasks.
These observations motivate jointly optimizing SD switching and agent-level scheduling rather than treating them as independent techniques.

However, predicting exact remaining execution time is difficult due to variable LLM generation and tool execution costs.
Instead, remaining agent steps provide a lightweight proxy for future LLM requests, which directly influence serving load.
Leveraging the statistical regularities and execution-state information discussed in~\autoref{sec:task_prediction}, remaining-step-aware scheduling can approximate the prioritization of shorter agent tasks without requiring precise execution-time prediction.

\subsection{Design Challenges}

The above observations motivate jointly optimizing agent-level scheduling and SD to address shifting performance bottlenecks and exploit their synergy.
Realizing this joint optimization raises three key challenges:

\begin{enumerate}[leftmargin=10pt]
    \item \textbf{Flexible and Low-Overhead SD Switching.}
    How can upper-layer policies control SD at runtime while minimizing switching and disabled-SD overhead?
    \item \textbf{Lightweight Agent Remaining-Step Prediction.}
    How can we estimate an agent task's remaining steps across diverse workflows and workloads using historical execution records and lightweight runtime information?
    \item \textbf{Bottleneck-Aware Joint Optimization.}
    How can agent-level scheduling and SD selection be coordinated to address shifting bottlenecks and maximize their combined benefits?
\end{enumerate}

\begin{figure*}[!t]
    \begin{center}
    \includegraphics[width=0.95\linewidth]{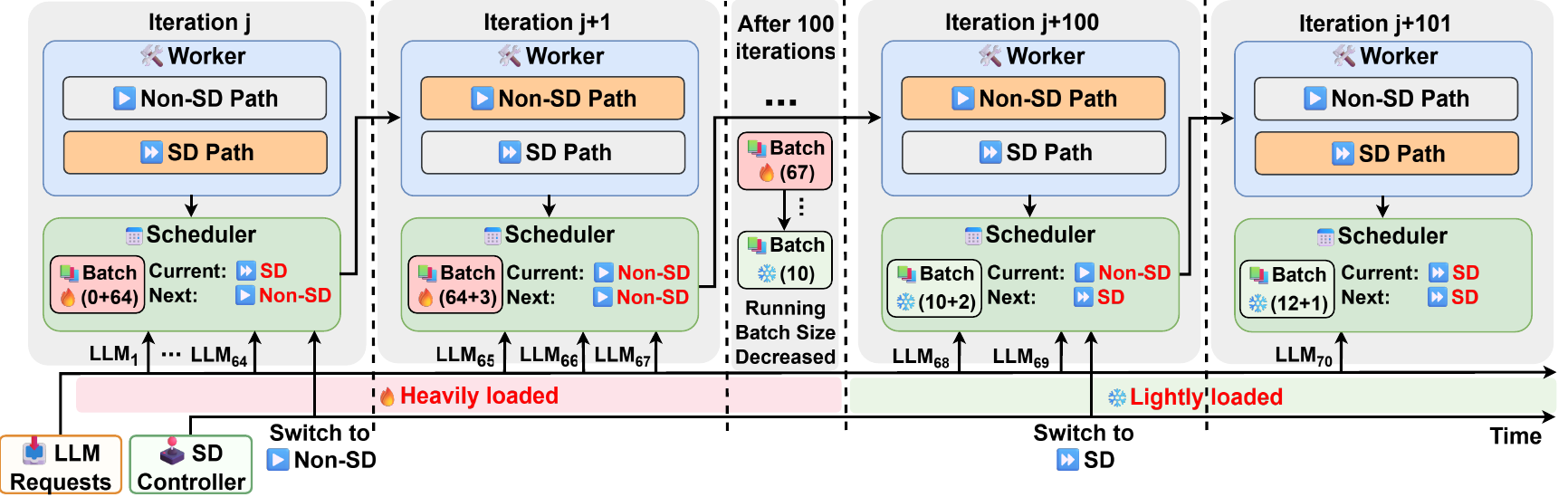}
    \end{center}
    \caption{
    Illustration of iteration-level SD switching in \sysname.
    At iteration $j$, heavy load triggers a switch to the non-SD path, effective at iteration $j+1$.
    As the batch size decreases, the controller switches back to the SD path at iteration $j+101$, while preserving compatibility with existing ASD policies.
    }
    \label{fig:sd_switch}
\end{figure*}

\section{Method}

To address the three challenges, \sysname integrates three key components.
First, externally controllable SD switching enables low-overhead transitions between native SD and non-SD paths.
Second, a lightweight history-based predictor estimates remaining agent steps to guide scheduling.
Finally, \sysname coordinates remaining-step-aware scheduling with load-adaptive SD selection, reducing high-load queueing delays while creating more SD opportunities under low load.

\subsection{Externally Controllable SD Switching}
\label{sec:4.1}

To address the limitations of existing ASD approaches discussed in~\autoref{sec:asd}, \sysname decouples SD switching policies from inference engine execution and enables iteration-level switching between native SD and non-SD paths, reducing non-SD overhead as shown in~\autoref{tab:adaptive_sd_comparison}.
As illustrated in~\autoref{fig:sd_switch}, this design consists of three components:

\noindent\textbf{External Control Interface.}
\sysname exposes a control interface that receives SD switching requests and forwards them to the scheduler to update the target execution path.
This design allows upper-layer switching policies to be customized or replaced without modifying the inference engine.

\noindent\textbf{Iteration-Level State Transition.}
Serving conditions can change while a continuous batch remains active, so coarse-grained reconfiguration may delay a needed switch.
\sysname therefore applies path changes at inference-iteration boundaries, allowing controller decisions to take effect at the next iteration without disrupting the current one.
The scheduler tracks the current and target paths, applying the requested transition at the next iteration.

\noindent\textbf{Path-Specific Worker Execution.}
\sysname prepares separate CUDA Graphs for native SD and non-SD paths during initialization, and the worker executes the corresponding path in each iteration.
Within the SD path, ASD policies can still dynamically adjust speculative decoding configurations, such as draft length.
Thus, \sysname is orthogonal to and compatible with existing ASD policies that operate within the SD path.

\subsection{Agent Remaining-Step Prediction}
\label{sec:4.2}

\begin{figure}[!b]
    \centering
    \includegraphics[width=\columnwidth]{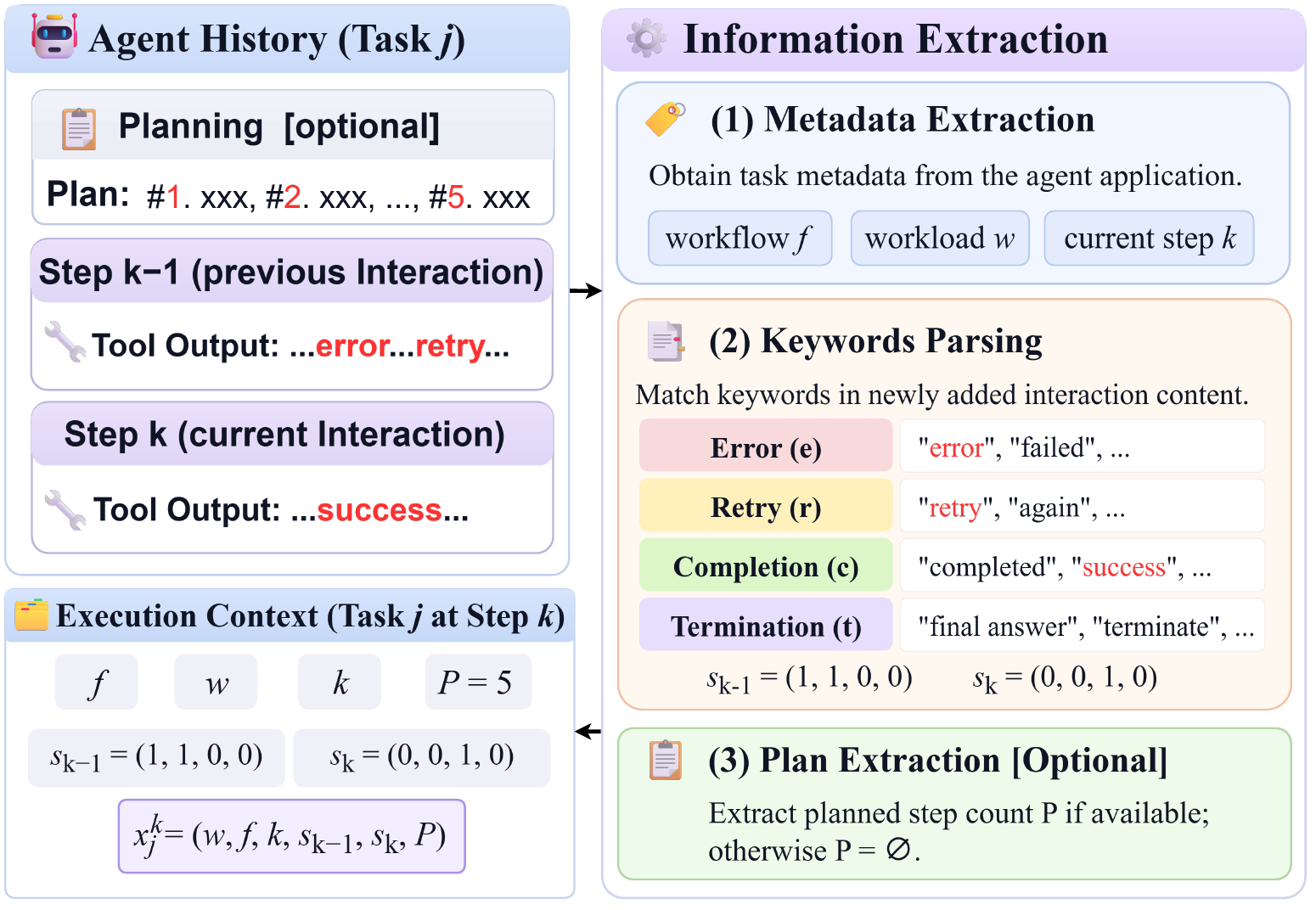}
    \caption{
    Illustration of request-level information extraction in \sysname.
    Task metadata, incremental execution states, and optional planning information form a unified execution context.
    }
    \label{fig:prompt_parser}
\end{figure}

Predicting agent remaining time is challenging due to variable LLM and tool execution costs under dynamic system conditions.
Instead, we estimate remaining steps as a proxy for future LLM requests, each of which contributes to the serving load regardless of its execution time.
Prioritizing tasks with fewer remaining steps thus helps reduce future batch sizes and creates more opportunities for SD acceleration.
To this end, \sysname develops a unified history-based estimator that combines historical execution patterns with runtime state information.

\noindent\textbf{Request-Level Information Extraction.}
As illustrated in~\autoref{fig:prompt_parser}, \sysname extracts three types of information for each agent task.
First, the agent application provides its workflow $f$, workload $w$, and current execution step $k$ as metadata.
Second, \sysname incrementally parses only the interaction content added since the previous LLM request, avoiding interference from historical prompts.
A lightweight keyword-based parser identifies four categories of signals: {\em error, retry, completion, and termination}, which are encoded into a four-dimensional binary state vector $s_k=(e_k,r_k,c_k,t_k)$.
Each dimension independently indicates the presence of its signal, allowing multiple signals to coexist.
The previous state $s_{k-1}$ is retained to capture recent execution transitions.
Third, when an explicit execution plan is available, \sysname extracts and retains its planned step count $P$; otherwise, $P$ is marked as unavailable.
Together, these features form a unified execution context $x_j^k=(w,f,k,s_{k-1},s_k,P)$ for task $j$ at step $k$.

\noindent\textbf{Offline History Profiling.}
\sysname collects the execution steps and per-step execution contexts of completed agent tasks.
These records are organized by workflow, workload, execution progress, and execution states to support statistical estimation.

\noindent\textbf{Unified Remaining-Step Estimation.}
When a plan is available, \sysname directly uses $P-k$ as the remaining-step estimate.
Otherwise, it estimates remaining steps for each agent task $j$ from historical tasks with similar execution contexts:
\begin{equation}\label{eq:remaining_steps}
\Phi_j =
\frac{1}{|\mathcal{H}(x_j^k)|}
\sum_{i \in \mathcal{H}(x_j^k)}
(N_i-k),
\end{equation}
where $\Phi_j$ denotes the predicted number of remaining steps of task $j$, $N_i$ is the total number of steps in historical task $i$, and $\mathcal{H}(x_j^k)$ contains matching historical tasks.
Historical records are matched hierarchically, first using $(w,f,k,s_{k-1},s_k)$, then falling back to $(w,f,k,s_k)$ and $(w,f,k)$ until a nonempty matching set is found.

\subsection{Load-Adaptive Agent Scheduling and SD Selection}
\label{sec:4.3}

\noindent\textbf{Optimization Objective.}
An agent task $R_i=\langle S_i^1,\ldots,S_i^{K_i}\rangle$ consists of $K_i$ dependent steps, each involving an LLM call followed by an optional tool execution.
Under a joint scheduling and SD selection policy $\pi$, its end-to-end JCT is
\begin{equation}
\begin{aligned}
\mathrm{JCT}_{\pi}(R_i)
=&\sum_{k=1}^{K_i}
\left(T_{\pi}^{\mathrm{Queue}}(S_i^k)
+T_{\pi}^{\mathrm{LLM}}(S_i^k)\right)\\
&+\sum_{k=1}^{K_i-1}T^{\mathrm{Tool}}(S_i^k),
\end{aligned}
\end{equation}
where $T^{\mathrm{Queue}}$, $T^{\mathrm{LLM}}$, and $T^{\mathrm{Tool}}$ denote queueing, LLM inference, and tool execution times, respectively.
Given a set of agent tasks $\mathcal{R}$, our objective is to minimize average JCT:
\begin{equation}
\pi^{*}=\operatorname*{arg\,min}_{\pi\in\Pi}
\frac{1}{|\mathcal{R}|}
\sum_{R_i\in\mathcal{R}}\mathrm{JCT}_{\pi}(R_i),
\end{equation}
where $\Pi$ denotes policies satisfying step dependencies and capacity constraints.
This optimization is challenging due to inter-step dependencies and resource contention; a related agent scheduling problem has been shown to be NP-hard~\citep{Astraea}.
Therefore, \sysname adopts heuristics that jointly prioritize agent tasks and adapt SD execution paths to serving load.

\noindent\textbf{Offline SD Speedup Profiling.}
For each model and hardware configuration, \sysname profiles SD speedup across different batch sizes before engine startup.
The break-even load $B^{*}$ is defined as the largest batch size at which SD achieves a speedup of at least 1.
SD is beneficial below $B^{*}$, while its overhead may degrade performance above this threshold.

\noindent\textbf{Future-Load Estimation.}
Since SD switching takes effect at the next inference iteration, the current running batch size alone may not reflect the future serving load.
\sysname therefore considers both the current running batch size $B_{\mathrm{running}}$ and the number of agent requests scheduled to resume from the router-proxy $B_{\mathrm{resume}}$:
\begin{equation}
\hat{B}=B_{\mathrm{running}}+B_{\mathrm{resume}}.
\end{equation}

\noindent\textbf{Load-Adaptive SD Selection.}
\sysname compares the estimated load $\hat{B}$ with the profiled threshold $B^{*}$ and selects the execution path through the external control interface:
\begin{equation}
\mathrm{Path}=
\begin{cases}
\mathrm{SD}, & \hat{B}\leq B^{*},\\
\mathrm{Non\mbox{-}SD}, & \hat{B}>B^{*}.
\end{cases}
\end{equation}
This enables SD acceleration under low load while avoiding its overhead under high load.

\noindent\textbf{Agent Task Scheduling.}
When agent tasks accumulate in the router-proxy queue under high load, \sysname prioritizes tasks with fewer predicted remaining steps:
\begin{equation}
j^{*}=\operatorname*{arg\,min}_{j\in\mathcal{Q}}\Phi_j,
\end{equation}
where $\mathcal{Q}$ denotes the queued tasks and $\Phi_j$ is estimated by the predictor (see~\autoref{eq:remaining_steps}).
By completing tasks with fewer LLM requests earlier, this strategy helps shorten high-batch periods and creates more opportunities for SD acceleration.

\begin{figure}[!tbp]
    \centering
    \includegraphics[width=\columnwidth]{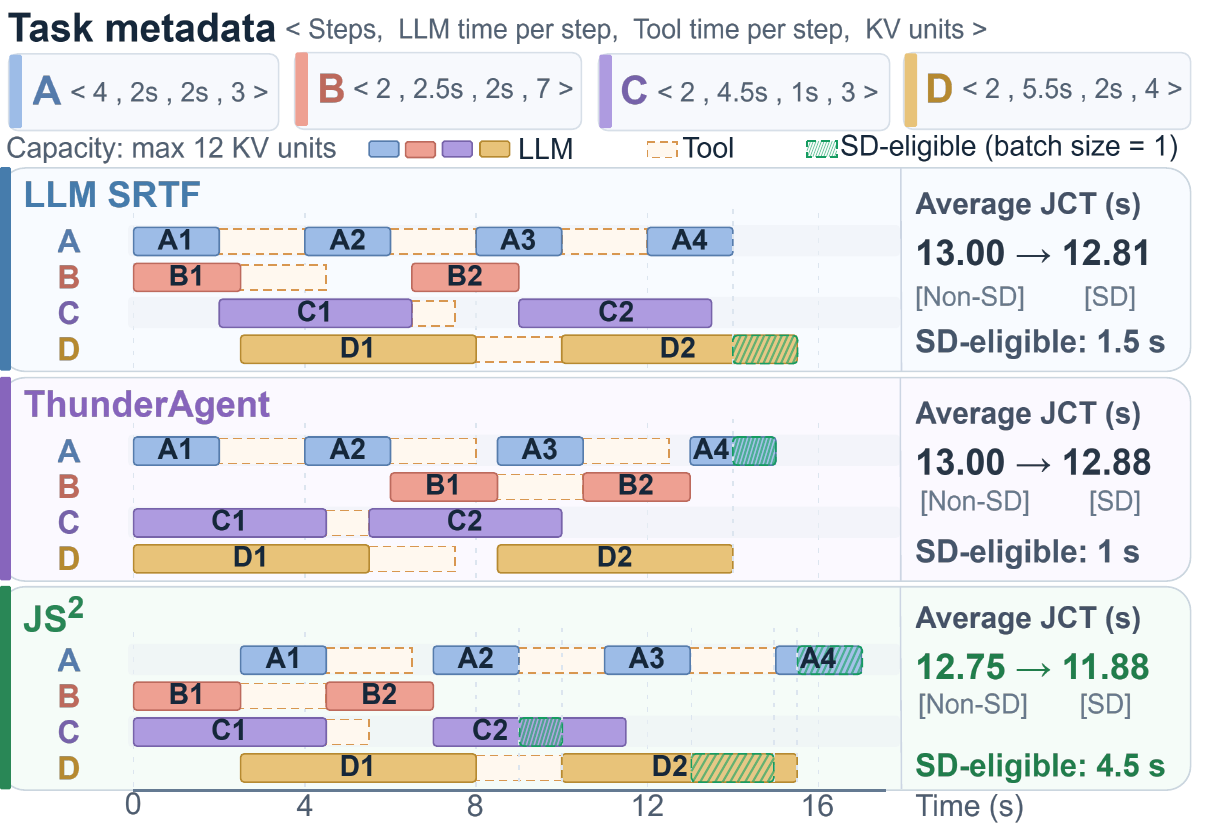}
    \caption{
        Illustrative scheduling comparison. \sysname achieves the lowest average JCT by increasing opportunities for SD acceleration. Green segments indicate SD-eligible intervals.
    }
    \label{fig:scheduling_pipeline}
\end{figure}

As shown in~\autoref{fig:scheduling_pipeline}, LLM-level SRTF prioritizes individual requests, while ThunderAgent favors concurrency.
In contrast, \sysname prioritizes tasks with fewer remaining steps, shortening high-batch periods and increasing SD-eligible execution from 1.0--1.5\,s to 4.5\,s.
Despite comparable JCT without SD, \sysname achieves the lowest average JCT (11.88\,s) with SD.
Although \sysname has a longer makespan in this finite example, continuous online arrivals and SD acceleration help mitigate tail delays.
Our experiments confirm comparable or lower tail latency (see~\autoref{fig:p95_throughput}).
\begin{figure*}[!t]
    \begin{center}
    \includegraphics[width=1.0\linewidth]{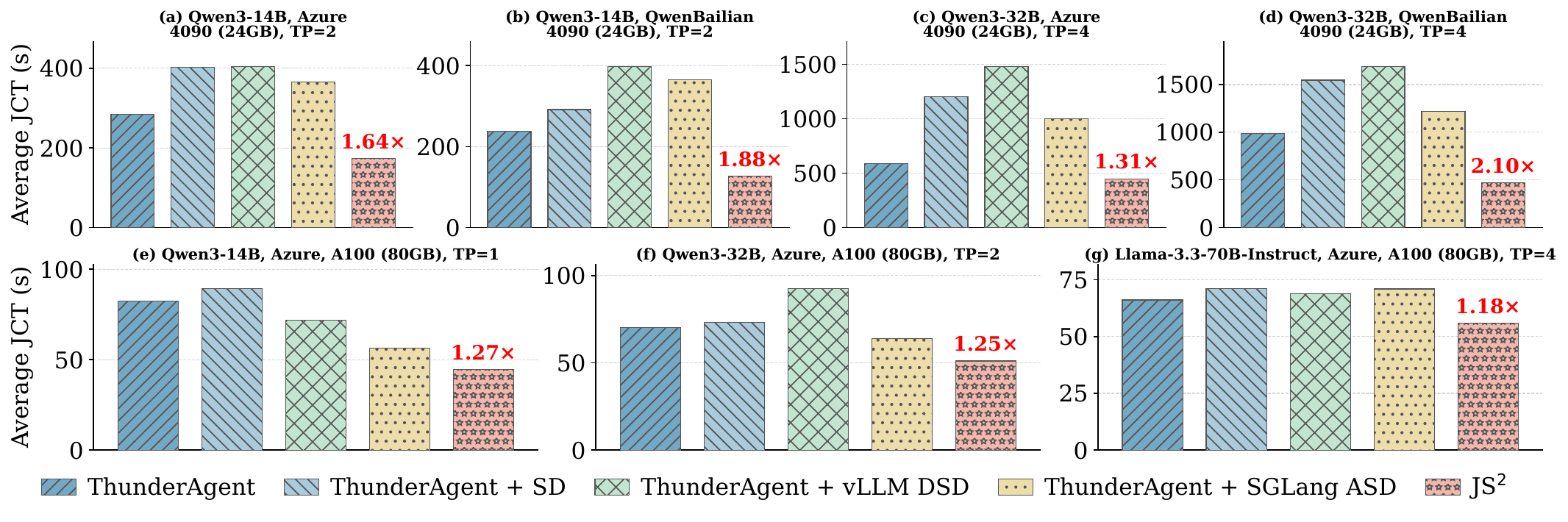}
    \end{center}
    \caption{
        Average end-to-end JCT across different serving configurations.
    }
    \label{fig:end2end_jct}
\end{figure*}

\section{Experiments}
\subsection{Experimental Setup}

\noindent\textbf{Hardware and Models.}
We evaluate \sysname on two 4-GPU servers: NVIDIA RTX 4090 (24GB, PCIe 4.0) with an Intel Xeon Gold 6338N CPU, and NVIDIA A100-SXM4 (80GB, NVLink 3.0) with an Intel Xeon Platinum 8369B CPU, both with 128 CPU cores.
We use Qwen3-14B, Qwen3-32B~\citep{Qwen3}, and Llama-3.3-70B-Instruct~\citep{Llama3}.
Detailed hardware resource configurations are provided in~\autoref{workload-and-resource-analysis}.

\noindent\textbf{Agent Workloads.}
We evaluate three workflows: ReAct~\citep{ReAct}, Plan-and-Solve~\citep{Plan-and-Solve}, and AgentFlow~\citep{AgentFlow}, across four datasets: MBPP~\citep{MBPP}, 2WikiMultihopQA~\citep{2wiki}, NL2Bash~\citep{nl2bash}, and Spider~\citep{spider}, covering coding, web search, Bash, and SQL tasks.
Detailed workload statistics are provided in~\autoref{workload-and-resource-analysis}.

\noindent\textbf{Online Traces.}
We use request arrival timestamps from two real-world traces: Qwen-Bailian~\citep{qwen-bailian} and Microsoft Azure~\citep{azure}.
Qwen-Bailian exhibits smoother load variation, whereas Azure shows more pronounced fluctuations (see details in~\autoref{online-trace-detail}).

\noindent\textbf{Baselines.}
We compare \sysname against ThunderAgent with native non-SD, fixed SD, vLLM Dynamic SD, and SGLang Adaptive SD configurations.

\noindent\textbf{Implementation.}
We implement \sysname by extending vLLM and SGLang with the runtime SD-switching interface (\autoref{sec:4.1}) and ThunderAgent's scheduler with remaining-step-aware scheduling (\autoref{sec:4.3}).

\noindent\textbf{Metrics.}
We evaluate average JCT, P95 JCT, and system throughput, with average JCT as the primary metric for end-to-end agent latency.

\subsection{End-to-End Performance}

\autoref{fig:end2end_jct} compares the average JCT of \sysname and four baselines across seven configurations. 
All speedups in this subsection are relative to the best baseline in each configuration.

\noindent\textbf{Different Request Traces.}
Under Azure's fluctuating load, \sysname adaptively switches between SD and non-SD paths, achieving \textbf{1.18--1.64$\times$} speedup across configurations.
Under Qwen-Bailian's sustained high load, SD offers fewer benefits and may increase JCT, making agent scheduling more critical.
In this setting, \sysname achieves \textbf{1.88$\times$} and \textbf{2.10$\times$} speedups on Qwen3-14B and Qwen3-32B, respectively.
Prioritizing tasks with fewer remaining steps also reduces subsequent batch sizes, creating more SD opportunities.

\noindent\textbf{Different Models and Hardware Platforms.}
On RTX 4090 GPUs, we deploy Qwen3-14B and Qwen3-32B with TP=2 and TP=4, respectively, representing resource-constrained settings with limited KV-cache capacity, concurrency, compute power, and inter-GPU bandwidth.
On A100 GPUs, we deploy Qwen3-14B, Qwen3-32B, and Llama-3.3-70B-Instruct with TP=1, TP=2, and TP=4, respectively, representing resource-abundant settings with larger KV-cache capacity, higher concurrency, stronger compute capability, and NVLink interconnects.
Detailed workload statistics and resource profiling results are provided in~\autoref{workload-and-resource-analysis}.
\begin{itemize}[leftmargin=10pt]
    \item
        Under resource-constrained settings, limited compute capability and memory resources introduce two challenges that \sysname addresses through its SD switching and completion-aware scheduling.
        \sysname achieves \textbf{1.31--2.10$\times$} speedup.
        \begin{itemize}[leftmargin=10pt]
            \item
                The system becomes compute-bound at small batch sizes, resulting in a low break-even point $B^{*}$ beyond which SD no longer provides acceleration.
                Consequently, non-SD (see ThunderAgent in~\autoref{fig:end2end_jct}) is preferred for a large fraction of execution.
                Existing dynamic SD baselines retain SD-related overhead and can perform worse than pure non-SD.
                In contrast, \sysname performs a runtime switch and incurs lower overhead in non-SD mode, reducing inference overhead and average JCT.
            \item
                Limited KV-cache capacity constrains the runnable batch size and causes more requests to remain queued.
                ThunderAgent optimizes spatial utilization by admitting as many requests as possible within the memory to maximize throughput, whereas \sysname considers the temporal dimension by prioritizing jobs with shorter predicted remaining steps.
                This allows \sysname to complete jobs earlier and reduce average JCT.
        \end{itemize}
    \item
        Under the resource-abundant setting, abundant KV-cache capacity reduces queueing, while its stronger compute capability increases the break-even load $B^{*}$ (defined in~\autoref{sec:4.3}), allowing SD to remain beneficial under higher serving loads.
        Consequently, there is less room for both scheduling and SD switching optimizations.
        Nevertheless, \sysname achieves the lowest average JCT, with \textbf{1.18--1.27$\times$} speedup, demonstrating its effectiveness even when the optimization headroom is limited.
\end{itemize}

\subsection{Tail Latency and Throughput}

\begin{figure}[!b]
    \begin{center}
    \includegraphics[width=1.0\linewidth]{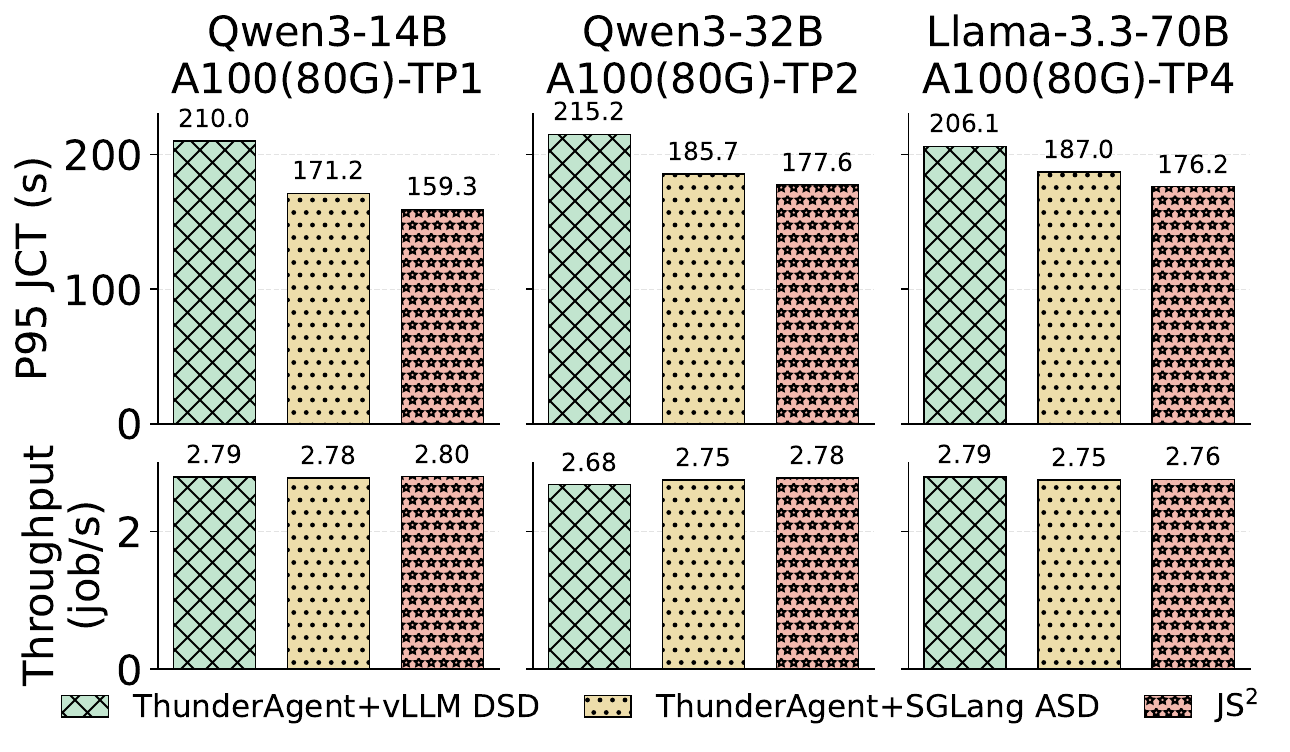}
    \end{center}
    \caption{
        P95 JCT and throughput on A100 GPUs.
    }
    \label{fig:p95_throughput}
\end{figure}

While \sysname reduces average JCT through remaining-step-aware scheduling, prioritizing shorter tasks may delay longer ones, potentially increasing tail latency.
However, under continuous online workloads, earlier completion of shorter tasks creates additional SD opportunities, which can also accelerate the decoding of long-running tasks and compensate for their additional queueing delays.
To evaluate this trade-off, we measure P95 JCT and system throughput across different models and tensor-parallel configurations.
As shown in~\autoref{fig:p95_throughput}, \sysname achieves comparable or lower P95 JCT than the baselines, indicating that its scheduling policy does not introduce additional tail-latency degradation.
Moreover, although \sysname prioritizes average JCT rather than maximizing throughput, it maintains comparable throughput across all evaluated configurations.
This suggests that prioritizing tasks based on remaining steps can improve completion efficiency without compromising overall resource utilization.

\subsection{Ablation Study}
We conduct an ablation study on the A100 platform using Qwen3-14B with TP=1 to evaluate the effectiveness of each component in \sysname, following the setup of~\autoref{fig:end2end_jct}(e).

\noindent\textbf{Remaining-Step Prediction Accuracy.}
Since our scheduler relies on queue ranking, we evaluate the predictor in~\autoref{sec:4.2} using Pair Accuracy and Top-10\% Recall, measuring pairwise ordering accuracy and recall of the true shortest 10\% of tasks, respectively.
As shown in~\autoref{tab:prediction_performance}, Plan-and-Solve achieves 100\% accuracy and recall by leveraging explicit planning information.
In contrast, AgentFlow and ReAct exhibit more dynamic execution patterns, making remaining-step prediction more challenging, although historical execution states still provide useful signals.
The Mixed setting jointly evaluates tasks from all three workflows.
Overall, \sysname achieves \textbf{84.1\% Pair Accuracy} and \textbf{82.2\% Top-10\% Recall}, providing reliable signals for scheduling.

\begin{figure}[!t]
    \centering
    \includegraphics[width=\linewidth]{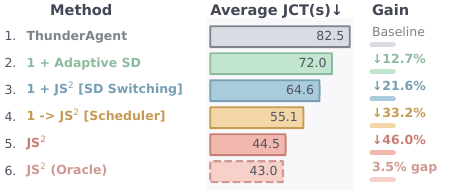}
    \caption{Ablation study of \sysname.}
    \label{fig:ablation}
\end{figure}

\noindent\textbf{Impact of Adaptive SD Mechanisms.}
We first compare different implementations of adaptive SD.
Replacing Adaptive SD with \sysname's SD switching reduces average JCT from 72.0\,s to 64.6\,s  (\autoref{fig:ablation}, 2 vs 3). 
This improvement mainly comes from \sysname's lower overhead in non-SD mode, while preserving comparable performance in SD mode.

\noindent\textbf{Impact of Scheduling.}
We next evaluate the impact of \sysname's completion-aware scheduling.
Compared with ThunderAgent (\autoref{fig:ablation}, 1), adding the \sysname scheduler (\autoref{fig:ablation}, 4) reduces average JCT from 82.5\,s to 55.1\,s, corresponding to a \textbf{1.50$\times$} speedup.
Compared with using only \sysname's SD switching mechanism (\autoref{fig:ablation}, 3), further incorporating \sysname's scheduler (\autoref{fig:ablation}, 5) reduces average JCT from 64.6\,s to 44.5\,s, corresponding to a \textbf{1.45$\times$} speedup.
This improvement demonstrates that remaining-step-aware scheduling provides complementary benefits, especially under high serving load where queueing becomes a major bottleneck, by prioritizing jobs with shorter predicted remaining steps to complete them earlier and reduce queueing delay.

\begin{table}[!t]
    \centering
    \caption{Remaining-step prediction performance.}
    \label{tab:prediction_performance}
    \renewcommand{\arraystretch}{1.5}

    \resizebox{\columnwidth}{!}{%
    \begin{tabular}{lcccc}
        \hline
        & \textbf{Plan-and-Solve}
        & \textbf{AgentFlow}
        & \textbf{ReAct}
        & \textbf{Mixed} \\
        \hline
        \#Items
        & 2167 & 3504 & 1611 & 7282 \\

        Pair Accuracy
        & 1.000 & 0.748 & 0.756 & 0.841 \\

        Top-10\% Recall
        & 1.000 & 0.630 & 0.673 & 0.822 \\
        \hline
    \end{tabular}
    }
\end{table}

\noindent\textbf{Gap to Oracle.}
Finally, we compare \sysname with an oracle configuration that uses oracle remaining-step information for scheduling.
The oracle configuration achieves an average JCT of 43.0\,s, compared with 44.5\,s for \sysname, leaving only a \textbf{3.5\%} gap (\autoref{fig:ablation}, 5 vs 6).
This small gap indicates that the predicted remaining steps are sufficient to achieve performance close to scheduling with ground-truth remaining steps.
\autoref{overhead-analysis} further demonstrates that the prediction and scheduling overheads are negligible.
\section{Conclusion}
In this work, we present \sysname, a bottleneck-aware serving system that jointly optimizes speculative decoding (SD) and agent-level scheduling to minimize end-to-end job completion time (JCT).
\sysname enables low-overhead, iteration-level switching between native SD and Non-SD execution paths and employs lightweight remaining-step prediction to guide agent scheduling.
By prioritizing tasks with fewer predicted remaining steps, \sysname reduces queueing delays and shortens high-batch periods, creating more opportunities for SD acceleration.
Together with load-adaptive SD selection, these mechanisms address shifting performance bottlenecks under dynamic serving loads.
Extensive evaluations across three LLMs, two GPU platforms, four agent workloads, three workflows, and two real-world serving traces demonstrate that \sysname achieves 1.18$\times$--2.10$\times$ speedups in average JCT over the best baselines, while maintaining comparable or lower P95 JCT and comparable throughput.

\section{Limitations and Future Work}

\noindent\textbf{Compatibility with More SD Methods.}
The current implementation of \sysname supports iteration-level switching for existing SD methods in vLLM and SGLang.
However, emerging SD methods, such as DFlash~\citep{DFlash} and DSpark~\citep{DSpark}, introduce different drafting and verification mechanisms.
Extending our switching mechanism to accommodate these methods is an important direction for future work, enabling broader compatibility and potentially greater acceleration benefits.

\noindent\textbf{Generalization of Agent Execution Prediction.}
Our remaining-step estimator relies on historical execution records, requiring sufficient data accumulation for new workflows and workloads.
Moreover, the current prediction and scheduling mechanisms primarily target sequential agent tasks, where LLM requests follow a serial execution pattern.
Future work will explore prediction with limited historical data and extend our approach to parallel agent workflows by incorporating task dependencies, execution graphs, and critical-path information.

\noindent\textbf{More Comprehensive SD Selection.}
The current SD selection policy primarily relies on near-future serving load, estimated from the running batch size and the number of requests scheduled to resume.
However, the effectiveness of SD also depends on workload characteristics, token acceptance rates, and drafting and verification costs.
Future work will incorporate additional signals, such as task-specific token acceptance rates and expected SD speedups, to enable more fine-grained and accurate switching decisions under dynamic serving conditions.

\clearpage
\bibliographystyle{IEEEtranN}
\bibliography{references}

\clearpage
\appendix

\section{LLM Inference Time Ratio}
\label{llm-time-ratio}

\begin{table}[H]
    \centering
    \caption{LLM inference time ratio across different agent tasks.}
    \label{tab:llm_time_breakdown}
    \begin{tabular*}{\columnwidth}{@{\extracolsep{\fill}}llc@{}}
        \toprule
        \textbf{Task} & \textbf{Dataset} & \textbf{LLM Time Ratio} \\
        \midrule
        \multirow{3}{*}{Web Search}
            & 2WikiMultihopQA & 60.46\% \\
            & Bamboogle~\citep{bamboogle} & 66.95\% \\
            & HotpotQA~\citep{hotpotqa} & 71.56\% \\
        \midrule
        Coding & MBPP & 99.87\% \\
        \bottomrule
    \end{tabular*}
\end{table}

We measure the fraction of end-to-end execution time spent on LLM inference across different agent tasks and datasets, as shown in~\autoref{tab:llm_time_breakdown}.
LLM inference accounts for the majority of the end-to-end latency across all evaluated settings, ranging from 60.46\% to 99.87\%.
Coding and Web Search tasks represent two contrasting cases.
In Coding tasks, tool execution mainly consists of running generated code and is usually very fast, whereas in Web Search tasks, agents often need to retrieve and access multiple webpages, making tool execution more sensitive to network latency.
Despite this difference, LLM inference remains the dominant latency component in both types of tasks.

As discussed in the motivation and illustrated in~\autoref{fig:motivation-a}, the dominant bottleneck within LLM inference shifts from decoding to queueing as concurrency increases, while prefill latency remains relatively small.
These observations motivate accelerating decoding under low serving loads and reducing queueing delays through scheduling under high serving loads.

\section{Impact of Batch Size on SD Speedup}
\label{sd-speedup}

\begin{figure}[!t]
    \begin{center}
    \includegraphics[width=1.0\linewidth]{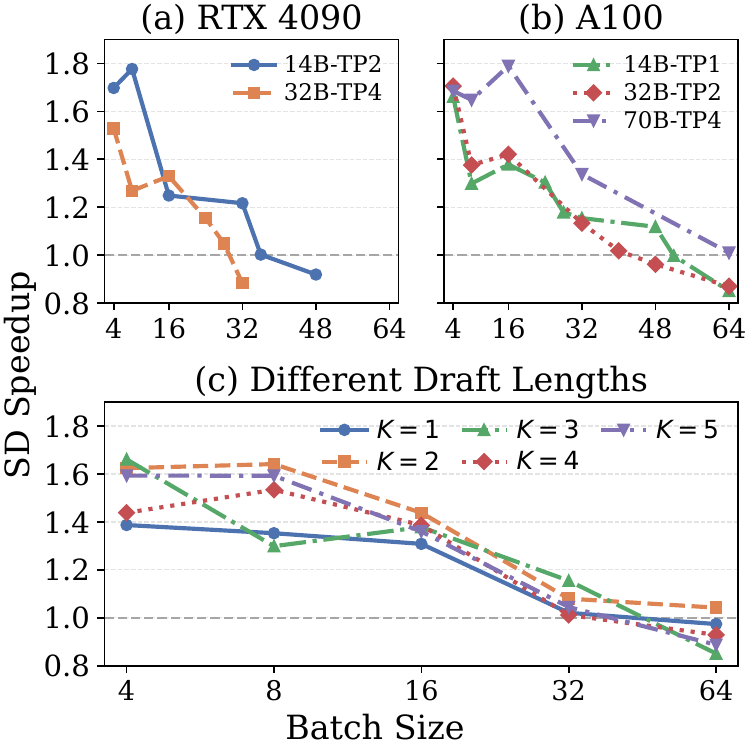}
    \end{center}
    \caption{SD speedup under different serving loads, model configurations, and draft lengths. Across both RTX 4090 and A100 GPUs, the benefit of SD generally decreases as batch size increases and can eventually become negative, while the optimal draft length also varies with serving load.}
    \label{fig:sd-speedup}
\end{figure}

To understand how serving load affects the effectiveness of speculative decoding, we measure the SD speedup across different batch sizes, GPU platforms, model configurations, and draft lengths, as shown in~\autoref{fig:sd-speedup}.
Across all evaluated configurations, the benefit of SD consistently decreases as the batch size increases and eventually becomes negative under sufficiently high serving loads.

Our experiments further reveal substantially different SD break-even points across hardware and model configurations.
Specifically, the break-even batch sizes are 36 for 4090-14B-TP2 and 28 for 4090-32B-TP4, compared with 52 for A100-14B-TP1, 40 for A100-32B-TP2, and 64 for A100-70B-TP4.
These results show that the serving load at which SD ceases to provide a performance benefit varies considerably across deployment configurations, highlighting the need to determine when SD should be enabled based on the specific serving environment.

We further observe that the speedup decreases more rapidly on RTX 4090 GPUs than on A100 GPUs.
In particular, A100 GPUs maintain positive SD speedup at larger batch sizes, resulting in a higher break-even batch size between SD and non-SD execution.
This observation suggests that the effective operating range of SD depends on the underlying hardware capability: more powerful GPUs can sustain the benefit of SD under higher serving loads, whereas less powerful GPUs enter the negative-speedup regime earlier.
Consequently, adaptive SD selection has greater optimization opportunities on resource-constrained GPUs, where choosing between SD and non-SD execution is more critical as the serving load changes.

Different draft lengths exhibit the same overall trend.
At low batch sizes, more aggressive configurations with longer draft lengths can achieve higher speedup, while increasing the batch size favors more conservative configurations with shorter draft lengths.
This load-dependent behavior motivates existing adaptive SD approaches to adjust the draft length according to the current serving load.
However, draft-length adaptation alone is insufficient at high loads: once the batch size becomes sufficiently large, even the most conservative configuration with a draft length of 1 can incur negative speedup.
Therefore, an effective adaptive SD mechanism should not only adapt the draft length, but also be able to completely disable SD and switch to the native non-SD execution path when speculative decoding is no longer beneficial.

\section{Execution Path Verification}
\label{execution-path-verification}

\begin{table}[!t]
    \centering
    \caption{Token consistency across adaptive SD approaches.}
    \label{tab:token_consistency}
    \footnotesize
    \setlength{\tabcolsep}{3pt}
    \begin{tabular*}{\columnwidth}{
        @{\extracolsep{\fill}}cccc@{}
    }
        \toprule
        \makecell{\textbf{Inference}\\\textbf{Engine}} &
        \textbf{Engine A} &
        \textbf{Engine B} &
        \makecell{\textbf{Token}\\\textbf{Consistency}} \\
        \midrule

        \multirow{4}{*}{vLLM}
        & \makecell{Dynamic SD ($\text{draft\_len}=0$)}
        & non-SD & $\times$ \\

        & \makecell{Dynamic SD ($\text{draft\_len}=0$)}
        & SD & $\checkmark$ \\

        & \sysname{} non-SD
        & non-SD & $\checkmark$ \\

        & \sysname{} SD
        & SD & $\checkmark$ \\

        \midrule

        \multirow{4}{*}{SGLang}
        & \makecell{Adaptive SD ($\text{draft\_len}=0$)}
        & non-SD & $\times$ \\

        & \makecell{Adaptive SD ($\text{draft\_len}=0$)}
        & SD & $\checkmark$ \\

        & \sysname{} non-SD
        & non-SD & $\checkmark$ \\

        & \sysname{} SD
        & SD & $\checkmark$ \\

        \bottomrule
    \end{tabular*}
\end{table}

Online inference engines typically employ different execution paths for SD and non-SD generation to improve efficiency, with different kernels used for key computations such as attention.
As a result, even under deterministic decoding settings (i.e., temperature $=0$ and top-$k=1$), SD and non-SD execution may generate slightly different token sequences for the same prompt due to differences in their underlying computation paths, although the generated contents are usually semantically similar.
Therefore, token-level consistency provides a practical indicator of whether two execution paths follow the same underlying computation.
We evaluate this behavior for vLLM Dynamic SD and SGLang Adaptive SD, as shown in~\autoref{tab:token_consistency}.
For both approaches, when SD is disabled by setting $\text{draft\_len}=0$, the generated tokens are inconsistent with those of the native non-SD engine but remain consistent with those produced by the native SD engine.
This indicates that setting $\text{draft\_len}=0$ does not fully switch execution back to the native non-SD path.
In contrast, \sysname's non-SD and SD paths produce tokens consistent with their respective native execution paths.
These results provide evidence that \sysname performs a complete runtime switch between the native SD and non-SD execution paths.

As discussed in the motivation, retaining the SD execution path even when $\text{draft\_len}=0$ introduces additional overhead compared with native non-SD execution.
In contrast, \sysname minimizes this overhead by directly switching to the native non-SD execution path.

\section{Online Serving Trace Analysis}
\label{online-trace-detail}

\begin{figure}[b]
    \begin{center}
    \includegraphics[width=1.0\linewidth]{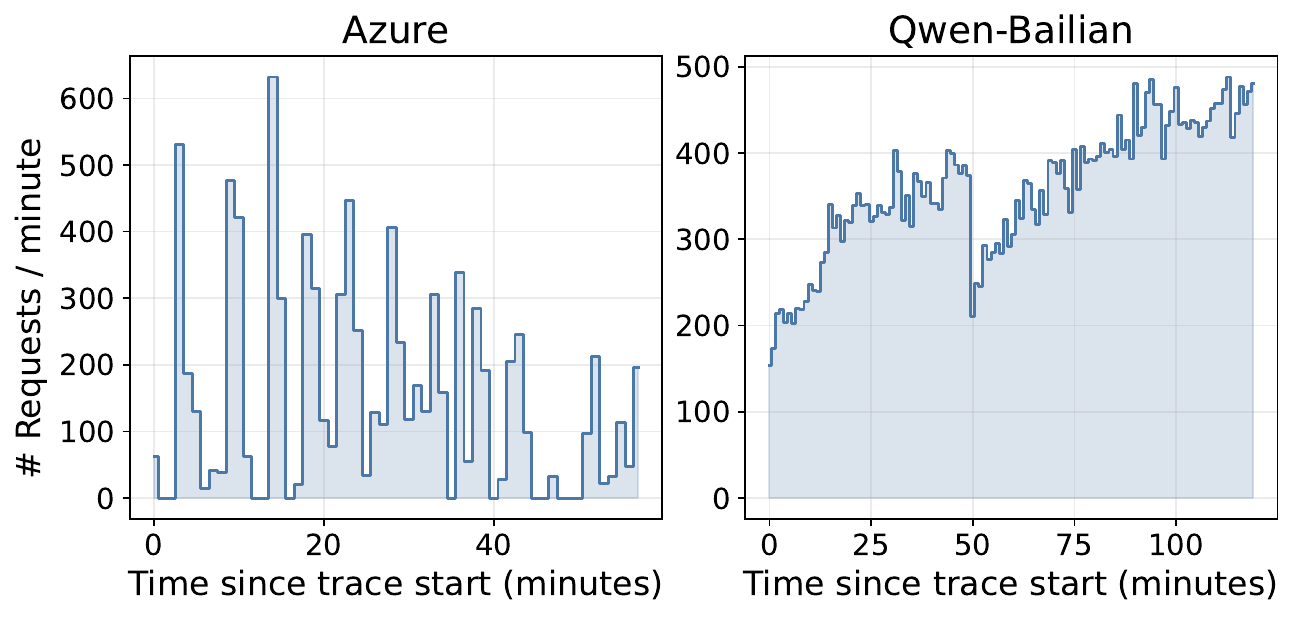}
    \end{center}
    \caption{
        Request arrival patterns of the Azure and Qwen-Bailian traces, measured as the number of requests per minute over time.
    }
    \label{fig:online_request_trace}
\end{figure}

We further examine the request-volume dynamics of the two online inference traces used in our end-to-end evaluation, as shown in~\autoref{fig:online_request_trace}, and summarize their statistics in~\autoref{tab:trace_statistics}.

\textbf{Azure Trace.}
The Azure trace exhibits highly bursty traffic, with frequent alternations between low-load periods and sharp traffic spikes.
Its average arrival rate is 152.05 requests/min, while the peak reaches 632 requests/min, about 4.16$\times$ the average.
Such rapid load fluctuations make SD beneficial during low-load periods, but may cause performance degradation when sudden bursts push the serving system into a high-load regime.
Therefore, this trace is well suited for evaluating whether adaptive SD methods can effectively adapt their SD decisions to rapidly changing serving loads.

\textbf{Qwen-Bailian Trace.}
In contrast, the Qwen-Bailian trace is considerably smoother and remains under sustained high load for most of its duration.
Its average arrival rate reaches 358.82 requests/min, more than twice the 152.05 requests/min observed in the Azure trace, while its peak is 488 requests/min.
Under such consistently high load, the benefit of SD is limited and can even cause performance degradation, while a large number of requests may accumulate in the waiting queue.
This trace is therefore well suited for evaluating both the non-SD overhead introduced by adaptive SD mechanisms under high load and the effectiveness of task scheduling strategies in reducing queueing delay.

\begin{table}[t]
    \centering
    \caption{Statistics of the Azure and Qwen-Bailian traces.}
    \label{tab:trace_statistics}
    \begin{tabular*}{\columnwidth}{@{\extracolsep{\fill}}lcc@{}}
        \toprule
        \textbf{Metric} & \textbf{Azure} & \textbf{Qwen-Bailian} \\
        \midrule
        \# Requests       & 8,819  & 43,058 \\
        Duration (min)    & 57.27  & 119.99 \\
        Avg. Req./min     & 152.05 & 358.82 \\
        Peak Min.         & 14     & 113    \\
        Peak Req.         & 632    & 488    \\
        \bottomrule
    \end{tabular*}
\end{table}

\section{Workload and Resource Analysis}
\label{workload-and-resource-analysis}

\begin{table}[b]
    \centering
    \caption{Agent workload statistics.}
    \label{tab:agent-workload-statistics}
    \footnotesize
    \setlength{\tabcolsep}{2.5pt}
    \begin{tabular*}{\columnwidth}{
        @{\extracolsep{\fill}}llccc@{}
    }
        \toprule
        \textbf{Workflow} &
        \textbf{Workload} &
        \textbf{\#Traces} &
        \makecell{\textbf{Avg. Prompt}\\\textbf{Tokens/Req.}} &
        \makecell{\textbf{Avg.}\\\textbf{Steps}} \\
        \midrule

        \multirow{2}{*}{AgentFlow}
            & Web Search & 100 & 1,502 & 12.84 \\
            & Coding     & 100 & 2,133 & 10.21 \\
        \midrule

        \multirow{2}{*}{\makecell[l]{Plan-and-Solve}}
            & Bash & 200 & 786 & 6.63 \\
            & SQL  & 200 & 627 & 6.21 \\
        \midrule

        \multirow{2}{*}{ReAct}
            & Bash & 200 & 2,293 & 3.32 \\
            & SQL  & 200 & 2,422 & 4.79 \\
        \midrule

        Overall & All & 1,000 & 1,504 & 6.49 \\
        \bottomrule
    \end{tabular*}
\end{table}

\textbf{Agent Request Analysis.}
We summarize the characteristics of the agent execution traces used in our experiments in~\autoref{tab:agent-workload-statistics}.
In total, the traces contain 1,000 execution traces across six workload--workflow combinations, with an average of 1,504 prompt tokens per request and 6.492 execution steps per task.
The request characteristics vary substantially across different combinations.
AgentFlow exhibits the largest average number of execution steps, reaching 12.84 steps on Web Search and 10.21 steps on Coding.
In contrast, ReAct requires only 3.315 and 4.785 steps on Bash and SQL respectively, while having relatively long prompts of more than 2,000 tokens per request.
Plan-and-Solve lies between these two workflows in terms of execution steps, while producing the shortest prompts among the evaluated combinations.
These statistics demonstrate the diversity of request lengths and execution patterns covered by our experimental workloads.

\begin{table}[t]
    \centering
    \caption{GPU memory usage and serving capacity.}
    \label{tab:serving-capacity}
    \footnotesize
    \setlength{\tabcolsep}{2.5pt}
    \begin{tabular*}{\columnwidth}{
        @{\extracolsep{\fill}}lcccc@{}
    }
        \toprule
        \textbf{Configuration} &
        \makecell{\textbf{Weights}\\\textbf{(GiB/GPU)}} &
        \makecell{\textbf{KV Cache}\\\textbf{(GiB/GPU)}} &
        \makecell{\textbf{KV}\\\textbf{Tokens}} &
        \makecell{\textbf{Supported}\\\textbf{Requests}} \\
        \midrule
        4090-14B-TP2 & 15.25 & 4.45  & 56,944  & 38  \\
        4090-32B-TP4 & 16.33 & 3.38  & 54,544  & 36  \\
        \midrule
        A100-14B-TP1 & 30.12 & 36.72 & 234,768 & 156 \\
        A100-32B-TP2 & 32.08 & 33.96 & 275,632 & 183 \\
        A100-70B-TP4 & 33.93 & 32.10 & 415,525 & 276 \\
        \bottomrule
    \end{tabular*}
\end{table}

\textbf{Hardware Resource Configuration Analysis.}
We further profile the inference-engine resource characteristics under the five experimental configurations, as shown in~\autoref{tab:serving-capacity}.
The number of supported requests is estimated by dividing the KV token capacity by the average prompt length reported in~\autoref{tab:agent-workload-statistics}.
On the RTX 4090 platform, the two configurations support only 36--38 concurrent requests, mainly due to the limited KV-cache capacity under the relatively tight GPU memory budget.
These configurations therefore represent a resource-constrained serving environment.
In contrast, the A100 configurations support approximately 156--276 requests, benefiting from their substantially larger GPU memory and KV-cache capacity, and thus represent a resource-abundant serving environment.
Overall, these measurements are consistent with our experimental design, where the RTX 4090 serves as a resource-constrained configuration, while the A100 serves as a resource-abundant configuration.

\section{Overhead Analysis}
\label{overhead-analysis}

To evaluate the runtime overhead introduced by \sysname's scheduling mechanism, we compare \sysname with ThunderAgent in terms of per-request processing overhead and paused-queue sorting overhead.
The former measures the additional cost of remaining-step prediction, while the latter evaluates the overhead of priority-based scheduling under different queue sizes.

\begin{table}[b]
    \centering
    \caption{Per-request scheduling overhead.}
    \label{tab:request-scheduling-overhead}
    \begin{tabular*}{\columnwidth}{
        @{\extracolsep{\fill}}lcc@{}
    }
        \toprule
        \textbf{Method} &
        \makecell{\textbf{Time}\\\textbf{(ms/request)}} &
        \makecell{\textbf{Increase}\\\textbf{(ms)}} \\
        \midrule
        ThunderAgent & 0.05086 & -- \\
        \sysname     & 0.13189 & 0.08103 \\
        \bottomrule
    \end{tabular*}
\end{table}

\autoref{tab:request-scheduling-overhead} compares the per-request scheduling overhead of ThunderAgent and \sysname.
ThunderAgent requires an average of \(0.05086\) ms per request, while \sysname requires \(0.13189\) ms because it additionally estimates the remaining execution steps.
The absolute increase is only \(0.08103\) ms per request.
Compared with agent inference and tool-execution latencies, which typically range from hundreds of milliseconds to several seconds, this additional overhead is negligible.

\begin{table}[t]
    \centering
    \caption{Paused-queue sorting overhead.}
    \label{tab:paused-queue-sorting-overhead}
    \begin{tabular*}{\columnwidth}{
        @{\extracolsep{\fill}}cccc@{}
    }
        \toprule
        \makecell{\textbf{Queue}\\\textbf{Size}} &
        \makecell{\textbf{ThunderAgent}\\\textbf{(ms)}} &
        \makecell{\textbf{\sysname}\\\textbf{(ms)}} &
        \makecell{\textbf{Increase}\\\textbf{(ms)}} \\
        \midrule
        10     & 0.0059 & 0.0084 & 0.0024 \\
        100    & 0.0482 & 0.0706 & 0.0224 \\
        1,000  & 0.4815 & 0.7855 & 0.3040 \\
        10,000 & 5.8395 & 9.0944 & 3.2549 \\
        \bottomrule
    \end{tabular*}
\end{table}

\autoref{tab:paused-queue-sorting-overhead} reports the sorting overhead under different paused-queue sizes.
\sysname is slightly more expensive than ThunderAgent because it prioritizes programs according to their predicted remaining steps.
With 1,000 paused programs, \sysname adds only \(0.304\) ms over ThunderAgent.
Even with 10,000 paused programs, the additional sorting time is approximately \(3.25\) ms.
Since ThunderAgent performs periodic scheduling at a configurable interval, this sorting overhead remains small relative to typical scheduling intervals.

\end{document}